\documentclass{article}

\usepackage[preprint, nonatbib]{neurips_2019}

\usepackage[utf8]{inputenc} % allow utf-8 input
\usepackage[T1]{fontenc}    % use 8-bit T1 fonts
\usepackage{hyperref}       % hyperlinks
\usepackage{url}            % simple URL typesetting
\usepackage{booktabs}       % professional-quality tables
\usepackage{amsfonts}       % blackboard math symbols
\usepackage{nicefrac}       % compact symbols for 1/2, etc.
\usepackage{microtype}      % microtypography

\usepackage{float}
\usepackage{graphicx}
\usepackage{mathtools}
\usepackage{wrapfig}
\usepackage{caption}

\title{Adaptive Transformers in RL}

\author{%
  \hspace{0.3cm} Jerrod Parker\thanks{Equal contribution, author ordering determined by the best return over 5 episodes of tic-tac-toe}\hspace{1cm}Shakti Kumar\footnotemark[1]\hspace{1cm}Panteha Naderian\\
  Department of Computer Science \\
  University of Toronto \\
  \texttt{\{jparker,shaktik,naderian\}@cs.toronto.edu} \\
}

\begin{document}

\maketitle

\begin{abstract}
Recent developments in Transformers have opened new interesting areas of research in partially observable reinforcement learning tasks. Results from late 2019 showed that Transformers are able to outperform LSTMs on both memory intense and reactive tasks. In this work we first partially replicate the results shown in Stabilizing Transformers in RL on both reactive and memory based environments. We then show performance improvement coupled with reduced computation when adding adaptive attention span to this Stable Transformer on a challenging DMLab30 environment. The code for all our experiments and models is available at \url{https://github.com/jerrodparker20/adaptive-transformers-in-rl}.
\end{abstract}

\section{Introduction}
Transformers have been shown to successfully model natural language through the use of self-attention to capture long range dependencies in text. It is natural to consider the extension of Transformers in Reinforcement Learning (RL) tasks which share a great deal with language modelling like the high correlation between adjoining states and the need to predict the future by knowing only the past. However, \cite{StableTransformers} demonstrated that when Transformers are directly applied to RL problems which require past memory, their performance is close to random. Thus the authors in \cite{StableTransformers} propose modifications to the TransformerXL (TXL) \cite{TXL} architecture leading to stable learning in RL tasks which require long range dependencies. They achieve results that surpass the previous benchmarks set by LSTMs.

The TXL architecture used in \cite{StableTransformers} restricts its memory block to small sizes in order to maintain computational feasibility. This means that deeper models are required to achieve a large context. The adaptive attention span \cite{AdaptiveAttention} solved this problem in language modeling by selectively attending to past timesteps which allows the use of much larger memory blocks. This is in contrast to the TXL where every attention head attends to the entire memory.

In this work we extend this adaptive attention span \cite{AdaptiveAttention} concept to partially observable reinforcement learning tasks. Our objective is to explore whether incorporating a variable length learnable attention span in each Transformer head allows the model to achieve similar or better performance to \cite{StableTransformers} with reduced computation. To our knowledge, this is one of the first works to explore the possibility of adaptive attention span in a reinforcement learning domain and reproduce results from \cite{StableTransformers}.

The rest of the paper is organized as follows. In Section 2 we discuss the background and some seminal work in imparting memory to RL agents. Section 3 discusses our Adaptive Transformer and the model design. We conduct extensive experiments to compare these methods on both a reactive environment which doesn't require past memory as well as a memory intense environment of DMLab30 \cite{DMLab}. The results from these experiments are summarized in Section 4. This is followed by a thorough discussion on our findings as well as open questions/limitations and possible future improvements.

\section{Related Work}
The original Transformer \cite{originalAttention} was able to achieve state of the art performance over LSTMs in 2017 by using a self-attention mechanism. The TransformerXL \cite{TXL} improved upon the original Transformer by allowing a much larger context size than was previously possible. This was achieved through maintaining a memory block which is similar to a hidden state in an RNN. Tokens could then attend to this memory block which encapsulates information from deep in the past.

Even with the breakthroughs in NLP, the use of Transformers in RL literature has been quite scarce. \cite{neuralAttentiveMetaLearner} attempted to use the original Transformer on simple tabular MDPs but was unable to achieve better than random performance. Stable Transformers \cite{StableTransformers} was the first to successfully apply Transformers to RL tasks. They again found that the standard Transformer architecture cannot achieve better than random performance on either MDPs or POMDPs. The authors were able to incorporate architectural modifications to the encoder blocks of the TXL \cite{TXL} to achieve state of the art performance on DMLab30 which consists of both reactive and memory based environments. Their major contribution was to place the layernorms \cite{LayerNorm} directly before the multi-head attention and feedforward submodules in the TXL block. This allowed the transformer block to act similar to an identity function early in training which they found to accelerate learning.

Since the introduction of TXL which was used as the base model in Stable Transformers, there have been two major developments in the Transformer architecture. Both have shown attractive reductions in computation and number of parameters compared to TXL while maintaining better performance in language modeling. The first of these improvements was adaptive attention span \cite{AdaptiveAttention} which uses past memory in a similar fashion to TXL. They learn the context length of each attention head and make it as short as possible while retaining performance. By doing this, it was shown that a much larger memory could be used without increasing computation which meant that long range dependencies were more easily learned. Another recent work \cite{PersistentMemory} removed the feed-forward layer in the Transformer block and augmented the self attention layers with a form of persistent memory whose purpose is similar to the feed-forward layer. The authors showed that this proposed architecture achieved similar performance to TXL on language modeling with close to half the number of parameters. We leave the exploration of persistent memory in RL to future work and have attached the code to do this for the interested reader.

Our main contribution is applying adaptive attention span to RL while maintaining the architecture proposed in Stable Transformer. Our hypothesis was that this will improve performance in memory intensive environments by allowing a much larger context length than TXL while using the same amount of computation.

\section{Model Architecture}
In each experiment our vision network was the large network from \cite{impala} which has 15 convolution layers. This was followed by an LSTM or Transformer. Linear layers were used at the output of the Transformer or LSTM for both the policy and value heads. Minor modifications were made to the IMPALA training pipeline to utilize the memory in both the Stable and Adaptive Transformers. Each large rollout from the actors was divided into smaller chunks to be processed sequentially by the Transformers with the previous chunks in the sequence becoming memory. The training pipeline is discussed thoroughly in Figure \ref{fig:modified_impala} of Appendix A.1.

% \begin{figure}
%     \centering
%     \includegraphics{maskfunction.png}
%     \caption{Caption}
%     \label{fig:mask_function}
% \end{figure}{}

\begin{figure}[H]
    \centering
    \includegraphics[scale=0.4]{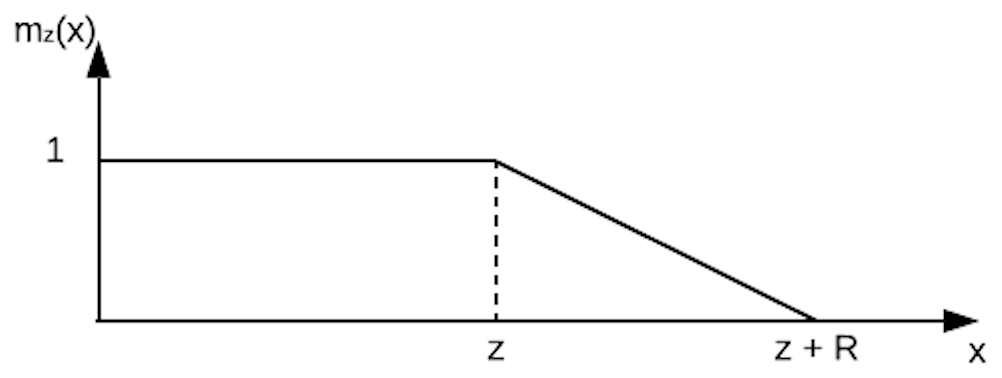}
    \caption{The mask $m_z$ as a function of span $z$}
    \label{fig:mask_function}
\end{figure}{}

\begin{wrapfigure}{Hr}{0.25\textwidth}
    \includegraphics[scale=0.35]{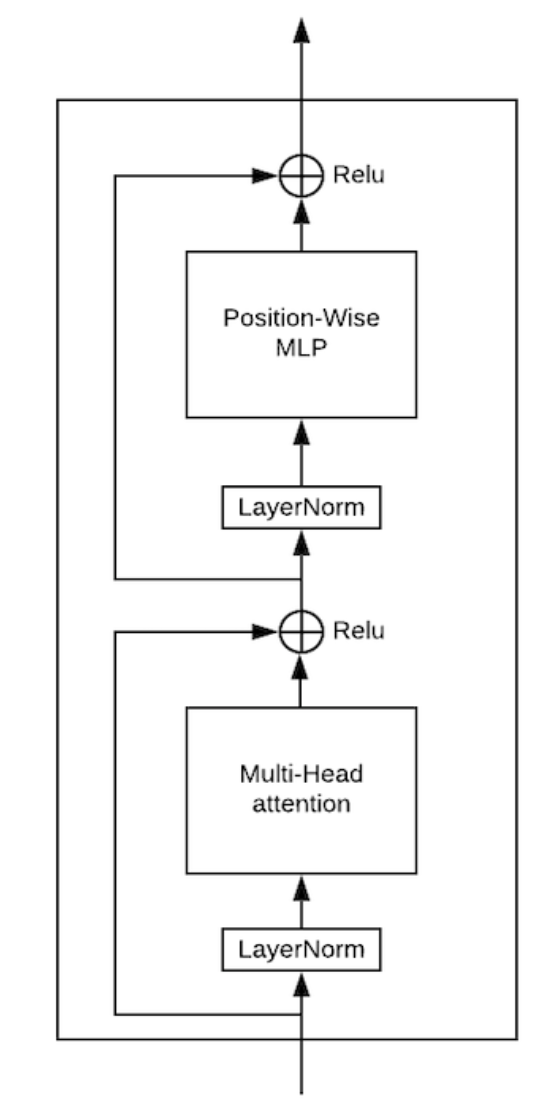}
    \caption{Stable Transformer decoder used in all our experiments.}
    \label{fig:transformer_decoder}
\end{wrapfigure}

We incorporate adaptive attention span into the Stable Transformer. This is done by maintaining one parameter $z$ per attention head which acts as a masking variable. A fixed parameter $R$ controls the smoothness of the mask. Figure \ref{fig:mask_function} depicts the masking function, and Equation \ref{eqn:firsteqn} shows the difference in how the attention weights are calculated between the original Transformer and when using adaptive attention span. In Equation \ref{eqn:firsteqn}, $a_{tr}$ is the  attention weight that token $t$ gives to token $r$ where $s$ denotes the attention logits. More information on the adaptive attention span model can be found in \cite{AdaptiveAttention}. Figure \ref{fig:transformer_decoder} shows the basic Transformer block in the Stable Transformer. The Adaptive version we use only changes the multihead attention submodule in this block so that it uses an adaptive attention span.

\begin{equation}
    a_{tr} = \frac{exp(s_{tr})}{\sum_{q=t-s}^{t-1}exp(s_{tq})}
    \;\;\;\;\;\;\xrightarrow[\text{to}]{\text{changes}} \;\;\;\;\;\;
    a_{tr} = \frac{m_z(t-r)exp(s_{tr})}{\sum_{q=t-s}^{t-1}m_z(t-q)exp(s_{tq})}
\label{eqn:firsteqn}
\end{equation}{}

When modeling a real world agent, it is desirable to be able to perform both memory intensive and reactive tasks. Hence in this section, we conduct experiments to compare the performance of a) LSTM, b) Stable Transformer and c) Stable Transformer using adaptive attention span on both a POMDP and MDP. For each of these experiments we used the IMPALA \cite{impala} actor-critic method with 32 actors each running on a separate CPU and a central learner on 2 NVIDIA Tesla P100 GPU with RMSProp as the optimizer. Also, the total number of parameters were kept similar among the models being compared. We first confirm results from \cite{StableTransformers} on a reactive task and then a memory intensive task. This is followed by a comparison of the Stable Transformer with and without adaptive attention span. Throughout this section, we refer to "Performance" as the final average 100 episode return achieved at 6M frames.

{}

\subsection{Stable Transformer on Reactive Environment}

One of the impressive findings from Stable Transformers was their ability to learn MDPs as effectively as LSTMs. They showed that their model could learn reactive tasks without sacrificing the number of layers which appears to stem from the ability of their variant to closely model the identity function early in training.

We ran experiments on Pong where we found that a 3.5M parameter, 3 layer Stable Transformer can learn it as quickly as a 1 layer with 1.8M parameters, while a 3.2M parameter 4 layer LSTM fails to perform better than random. The results are included in the center plot of Figure \ref{fig:main_plots} which show that both 1 and 3 layer Stable Transformers start converging within 3M steps with a last 100 average episode return of greater than 15. A 1.5M parameter 1 layer LSTM starts converging late around 5M steps while a 4 layer LSTM completely fails to do so. A learning rate of 0.0004 was found to be the best for both architectures on Pong. Other hyperparameters used for this experiment are listed in Tables \ref{tab:common_params} and \ref{tab:stable_params} of Appendix A.4.

These results show the attractive property of Stable Transformers to be able to learn reactive tasks even with deeper architectures which appears to be in contrast to LSTMs.

\subsection{Stable Transformer on Memory Based Environment}

The main reason for introducing Transformers in RL is for modeling dependencies in POMDPs. \cite{StableTransformers} showed the ability of the Stable Transformer to learn memory intensive tasks such as “Numpad” \cite{Numpad}  much better than LSTMs. To partially replicate this finding, we compare the performance of the Stable Transformer and LSTM on \texttt{rooms\_select\_nonmatching\_object} from DMLab. In this game, the agent is placed in a room which consists of a teleportation pad and another object. The goal is to use the teleportation pad and then choose the object in the resulting room which does not match the object from the first room. It is quite evident that the agent needs to use past information to choose the proper object when in the second room.

We compared a 4 layer LSTM to a 3 layer Stable Transformer with comparable number of parameters at 3.2M and 3.5M respectively. The learning rates used were the ones that achieved the best performance during our hyperparameter sweep summarized in Figures \ref{fig:stableDifferentLrsPlot} and \ref{fig:lstmDifferentLrsPlot}. The result summarized in the left plot of Figure \ref{fig:main_plots} shows that the Stable Transformer matches the performance of the LSTM over 6M environment interactions. This is as expected since the results from \cite{StableTransformers} on DMLab gave comparable performance between the LSTM and Stable Transformer for the first 2B frames beyond which the Stable Transformer starts to outperform the LSTM. The hyperparameters used for this experiment can be found in Tables \ref{tab:common_params} and \ref{tab:stable_params} of the Appendix A.4.

\subsection{Using Adaptive Attention Span in the Stable Transformer}

The size of a TXL memory is constrained by the fact that every attention module attends to the entire memory. Adaptive attention span has the benefit of being able to use a larger memory when only few layers have large attention spans which allows the average span to stay relatively small. This large memory comes without the computational burden that a TXL would have when using a memory of the same size. Hence we next compare the Stable Transformer with its adaptive attention span counterpart on DMLab’s memory intense \texttt{rooms\_select\_nonmatching\_object}.

%REPLACED THIS ABOVE Adaptive attention span is beneficial when different layers have differing attention spans but the overall average span is smaller than the memory length of TXL. This means that a larger memory size can be used than in TXL while maintaining the same computation.

\textbf{Where should Adaptive have an advantage over TXL in this game?} \\
% \textbf{Why should adaptive perform better than Stable Transformer in this game?}\\
We expected attention heads at higher layers to start capturing global context which is feasible once there are better representations of the game states at different times from the early layers. This is supported by the results in \cite{AdaptiveAttention} on language modelling which showed that the early layers learned spans of under 50 tokens while the higher layers learned spans of much larger than 1000 tokens.
In our experiment we expected that the early layers would capture local information such as which room the agent is in because it should only need to attend to a few states in the past to figure this out by confirming the existence of a teleportation pad.
Local information should also help the agent understand its current position in the room relative to the pads which can be necessary to make the next move as shown in Figure \ref{fig:non_matching_object}. When the agent teleports to the second room it needs to attend further in the past to know what object it was shown in the first room. Hence we expected some layers to have large spans to capture global information with the others having smaller spans to capture local information. If this is the case, the model may be able to avoid
enough compute in the smaller span layers that it gains computational advantages over TXL.

\begin{figure}
    % \centering
    \includegraphics[scale=0.2]{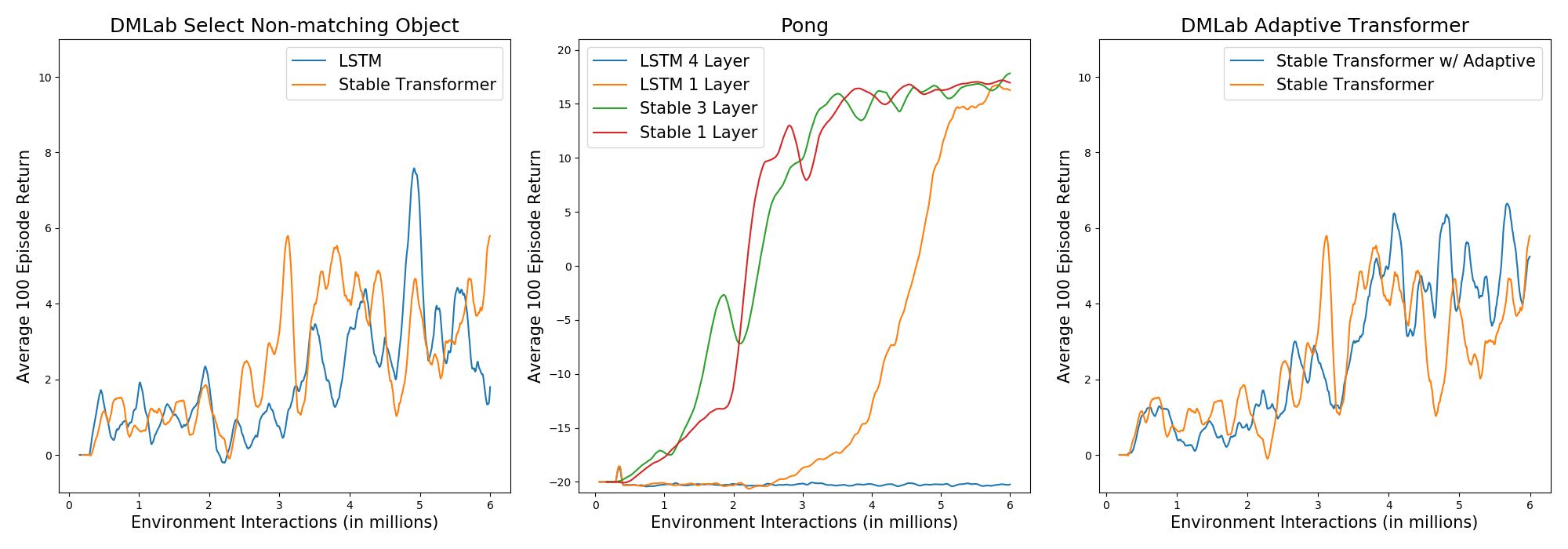}
    \caption{\textbf{Left}: The Stable Transformer vs LSTM using the best hyperparameters found based on average 100 episode return after 6M environment interactions. \textbf{Center}: The average 100 episode return of the LSTM and Stable Transformer when varying the number of layers in each in Pong. \textbf{Right}: The Stable Transformer with and without adaptive attention span on the \texttt{rooms\_select\_nonmatching\_object} environment.}
    \label{fig:main_plots}
\end{figure}{}

\textbf{Experiment Setup} Since the loss dynamics between adaptive and non-adaptive versions match very closely, we believe that both the models should use similar learning rates. Hence we chose to use the best performing learning rate from the Stable Transformer for both models in this experiment.

We set the initial memory length of the Adaptive version to 400 which is twice as large as that of Stable Transformer. This was done under the assumption that the model will be able to learn the smallest required attention span in each layer allowing the use of a large memory while keeping computation comparable with TXL. We then continually increased the L1 regularization coefficient on the spans until the performance stopped increasing which is shown in Figure \ref{fig:adaptiveAppendixPlots}.

\begin{wrapfigure}{Hr}{0.45\textwidth}
    \includegraphics[scale=0.4]{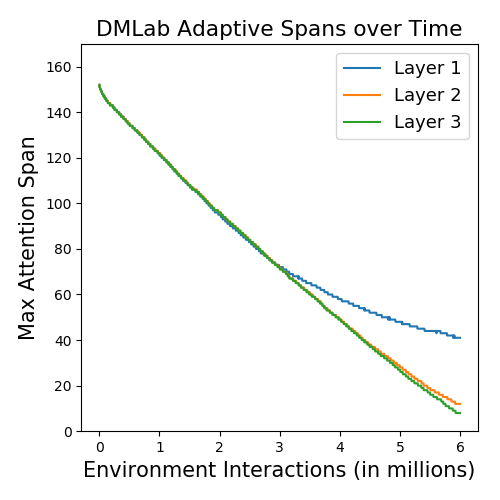}
    \caption{The maximum spans for the attention heads in each layer of the Adaptive model with penalty of 0.025. Span values for other penalties are listed in Figure \ref{fig:adaptiveAppendixPlots}.}
    \label{fig:span_lengths_main}
\end{wrapfigure}{}

\textbf{Results} Based on the right plot in Figure \ref{fig:main_plots} we can see that the 3 layer Adaptive version with 3.3M parameters outperforms the original 3 layer Stable Transformer with 3.5M parameters and shows a consistent increase in the average 100 episode return while maintaining a smaller variance as well. The maximum attention spans learned with the final span penalty value of 0.025 were 33, 2 and 2 in the first, second and third layers respectively as shown in Figure \ref{fig:span_lengths_main}. These values are in partial agreement with our hypothesis that a few layers should have large contexts and the others should learn smaller ones, though we suspect that the reason for spans being so low in the second and third layers is due to those layers not being needed in this game at all. The minuscule spans learned in the two higher layers provide large computational savings otherwise impossible in TXL architectures.

% INCLUDE THIS IN THE RESULTS
% The attention spans per layer are shown in Figure \ref{fig:span_lengths_main} and agree with this assumption.

Another interesting corollary of the small attention spans learned in the second and third layers was that using this penalty, the model learned how many attention layers were actually necessary. One of the very attractive properties of this form of regularization is that it helps avoid computing some attention logits completely compared to other regularizers like L2 in an MLP where all logits still need to be calculated.

We give an exhaustive comparison of all the architectures used in Table \ref{tab:return_table}. For each model, we did a hyperparameter sweep and found the set that gave the best final average return over 100 episodes after 6M steps.

\begin{table}[ht]
    \centering
    \captionsetup{justification=centering}
    \begin{tabular}{l*{3}{c}r}
    \textbf{Architecture} & \textbf{\#Parameters}
    & \textbf{Reactive}  & \textbf{Memory}\\
    \hline
    Stable (3 layer)
    &  3.5M & 17.62 & 7.17\\
    Adaptive (3 layer)
    &  3.3M & --- & 8.84 \\
    LSTM (4 layer)
    &  3.2M & -20.16 & 5.94 \\
    Stable (1 layer)
    & 1.8M & 16.92 & ---\\
    LSTM (1 layer)
    & 1.5M & 16.56 & --- \\ \\
    \end{tabular}
    \caption{Final 100 episode average returns of all the different architectures. Reactive: Atari Pong, \\ Memory: DMLab30's \texttt{rooms\_select\_nonmatching\_object} }
    \label{tab:return_table}
\end{table}{}

\section{Limitations}
Due to compute limitations, we had to trim down the number of layers used in Stable Transformers \cite{StableTransformers} from 12 to 3. This limitation also prevented us from doing an extensive hyperparameter sweep so we borrowed most of our parameters from those used in IMPALA \cite{impala}. It is expected that the convergence of the discussed models can be improved with a more extensive search on hyperparameters like warmup steps and decay rate for the scheduler. Furthermore, we have demonstrated our results on only 2 environments. Hence a more extensive experimentation is required to fully replicate the results from \cite{StableTransformers}. We also reported the performance of our models for only 6 million frames (approximately 4 hours of runtime on 32 CPUs and 2 P100 GPUs) as compared to 1 billion frames reported in IMPALA \cite{impala} and 10 billion in Stable Transformers. Even though our results closely agreed with \cite{StableTransformers}, we think that running the experiments for a greater number of frames will give a more fair comparison among the models shown.

Future work should attempt to compare Stable Transformers with and without adaptive attention span on environments like “Numpad” and “Memory Maze” from \cite{StableTransformers} which require much larger context than the DMLab environment we used. We expect this to amplify the performance benefits we found when using adaptive span in the Stable Transformers. Two additional future experiments are to use persistent memory \cite{PersistentMemory}  in the Stable Transformer as well as GRU gating as was done in \cite{StableTransformers}. We include the code to easily use both of these features.

\section{Conclusion}

In this project, we partially replicated results from \cite{StableTransformers} on both reactive and memory based environments. We showed that adaptive attention span can be successfully applied to partially observable tasks in RL and performs better than the Stable Transformer. Our experiments on DMLab30's \texttt{rooms\_select\_nonmatching\_object} and Atari's Pong showed more stable learning with higher final average rewards when using the Stable Transformer with adaptive attention spans. This shows that when Transformers are used in RL, the use of adaptive attention span likely improves performance and reduces computation. We look forward to exploring the use of adaptive attention span in RL on a more extensive set of environments in the future.

\section*{References}

\bibliographystyle{unsrt}
\bibliography{dissertationbib}

\newpage
\appendix
\section{Appendix}
\subsection{Training Pipeline}
Each actor does a rollout of length unroll\_length frames and pushes these trajectories into one of forty buffers. The unroll\_length was set to 400 for \texttt{rooms\_select\_nonmatching\_object} and 240 for Pong. The learner model which runs on 2 NVIDIA Tesla P100 GPUs divides these trajectories into chunks of size mini\_batch (which equals 100 for \texttt{rooms\_select\_nonmatching\_object} and 80 for Pong). Each chunk is concatenated with memory of size mem\_len which is the hidden state from the Transformer from the previous chunk and is then fed into the learner. The mem\_len was set to 100 and 200 for Pong and \texttt{rooms\_select\_nonmatching\_object} respectively in Stable Transformer and 400 for Adaptive in \texttt{rooms\_select\_nonmatching\_object}. The learner calculates vtrace returns and backpropagates on the sum of 3 loss functions a) policy gradient loss, b) baseline loss and c) entropy loss as discussed in \cite{impala}. The learner updates the parameters of the actors after each batch of trajectories is backpropagated and learnt.

For experiments using Stable Transformer with adaptive attention span, the span parameter $z$ was learned by penalizing the L1 norm of the current span values.

\begin{figure}[H]
    \centering
    \includegraphics[scale=0.7]{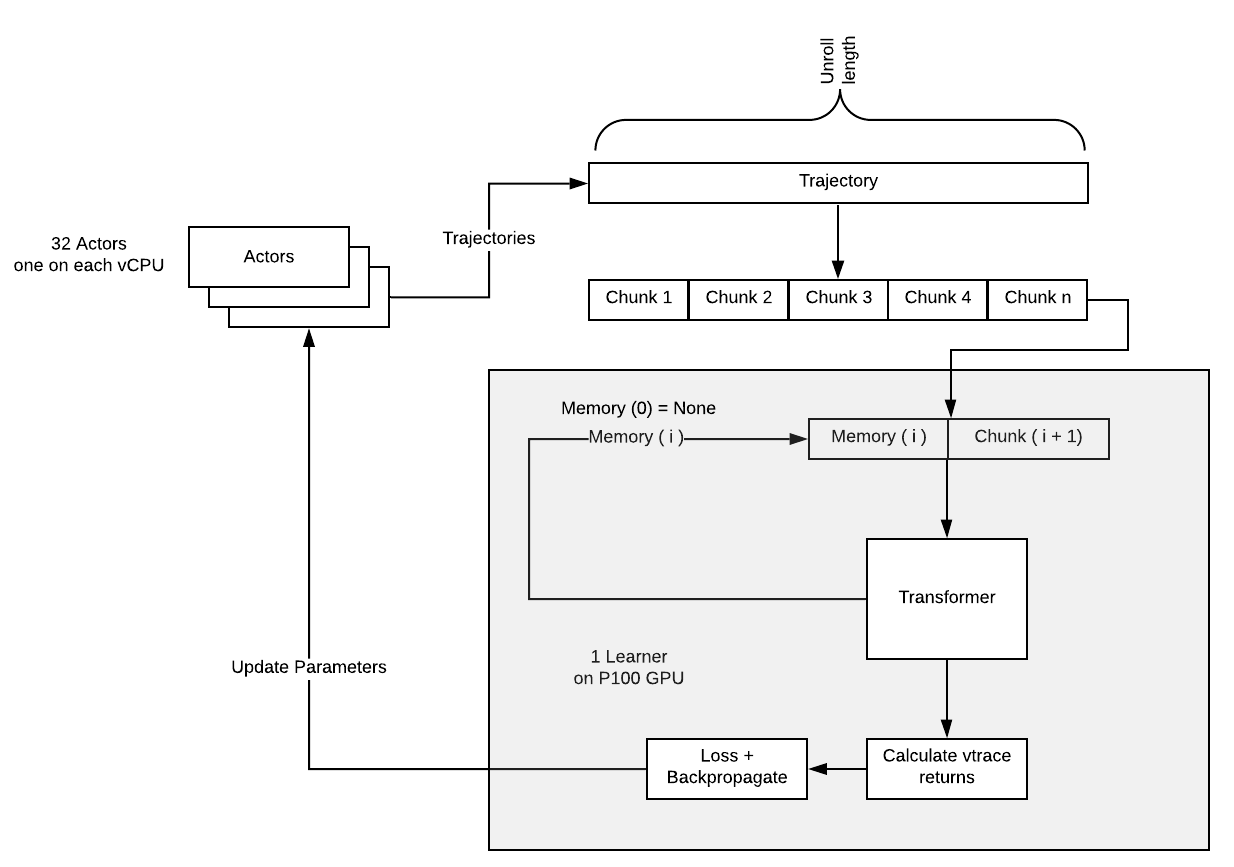}
    \caption{Modified IMPALA used in our experiments}
    \label{fig:modified_impala}
\end{figure}{}

\newpage
\subsection{Additional Experiments}
\begin{figure}[H]
    \centering
    \includegraphics[scale=0.22]{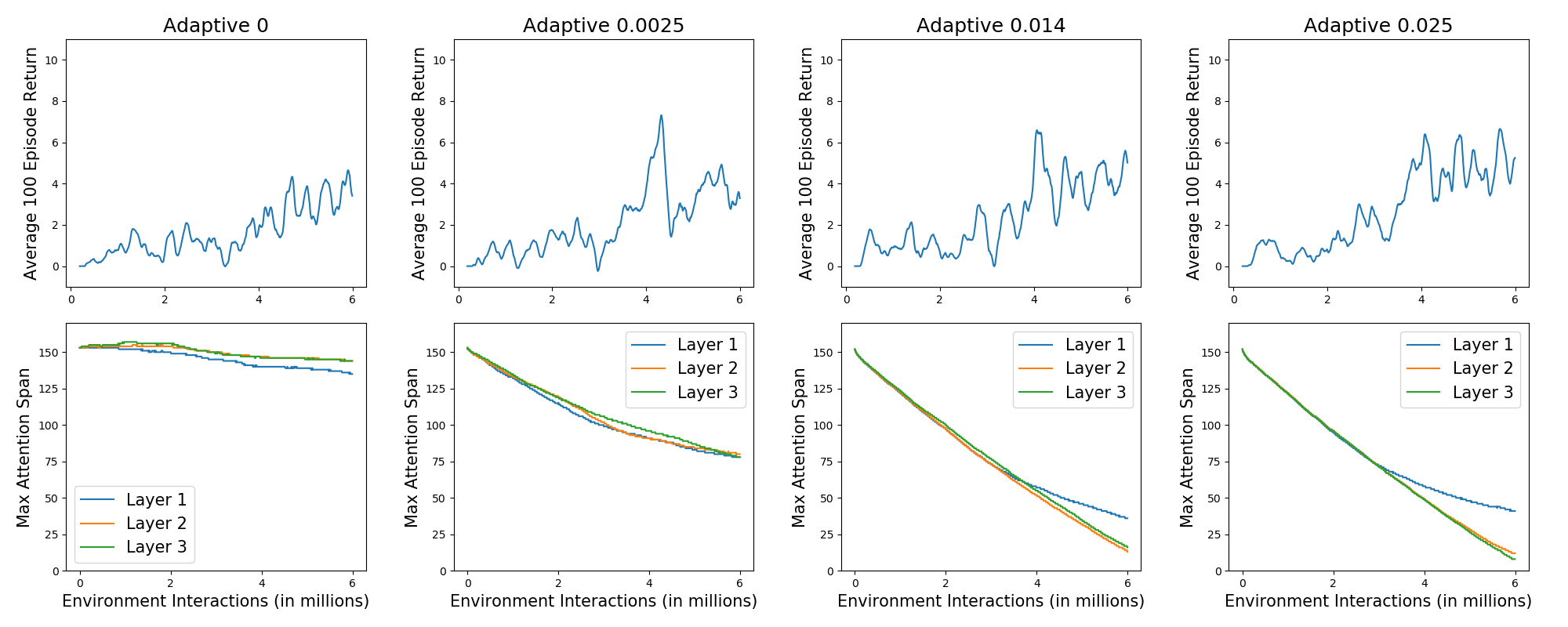}
    \caption{
    \textbf{Top:} Plots of the returns of the Stable Transformer using adaptive attention span on \texttt{rooms\_select\_nonmatching\_object} with different penalty coefficients for the span parameters as given in the titles. \textbf{Bottom:} Maximum attention span for each layer over the course of training.
    }
    \label{fig:adaptiveAppendixPlots}
\end{figure}{}

\begin{figure}[H]
    \centering
    \includegraphics[scale=0.3]{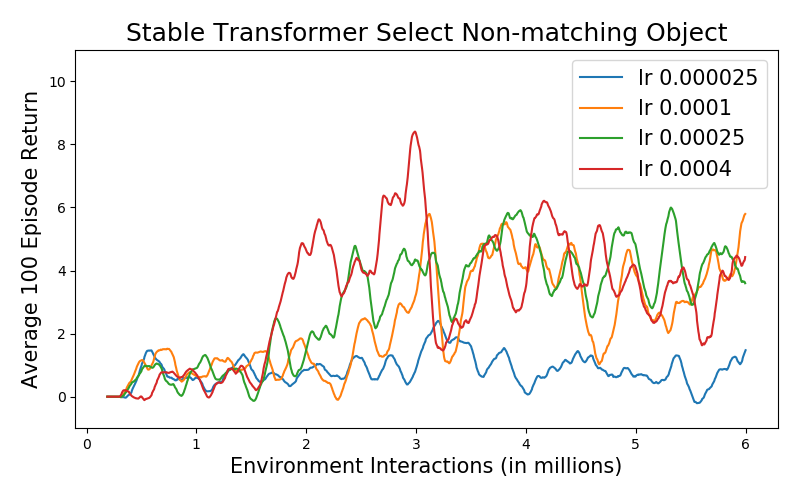}
    \caption{
    Plots of returns on \texttt{rooms\_select\_nonmatching\_object} using the Stable Transformer with different learning rates (lr) for RMSProp.
    }
    \label{fig:stableDifferentLrsPlot}
\end{figure}{}

\begin{figure}[H]
    \centering
    \includegraphics[scale=0.3]{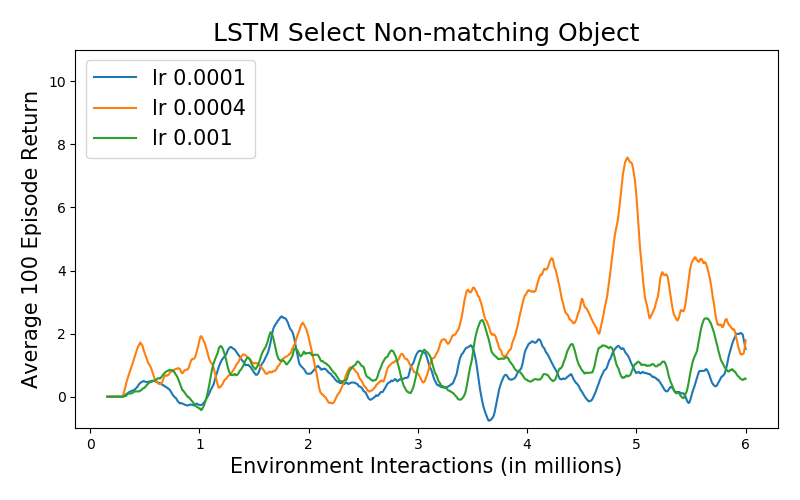}
    \caption{
    Plots of returns on \texttt{rooms\_select\_nonmatching\_object} using the LSTM transformer with different learning rates (lr) for RMSProp.
    }
    \label{fig:lstmDifferentLrsPlot}
\end{figure}{}

\newpage

\subsection{DMLab30 \texttt{rooms\_select\_nonmatching\_object}}
\begin{figure}[H]
    \centering
    \includegraphics{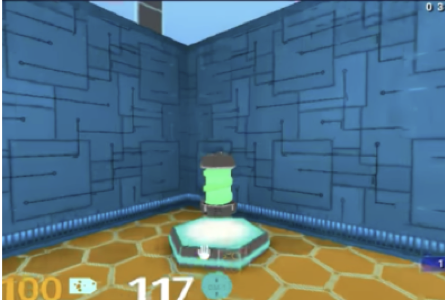}
    \caption{This game state is an example of the partially observable nature of \texttt{rooms\_select\_nonmatching\_object}. Here the agent can not know what room they are in or where they are in that room based on this image alone. For example, if the agent is in the first room, they cannot tell whether the teleportation pad is to our left or right.}
    \label{fig:non_matching_object}
\end{figure}{}

\subsection{Common Hyperparameters}
\begin{table}[ht]
    \centering
    \begin{tabular}{l*{3}{c}r}
    \textbf{Parameter}
    & \textbf{Value}  \\
    \hline
    Total steps
    &  6 million \\
    Batch size
    &  16 \\
    No. of actors
    &  32 \\
    No. of buffers
    &  40 \\
    No. of learner threads
    & 1 \\
    Action repeats
    & 1 \\
    Entropy cost
    & 0.01 \\
    Baseline cost
    & 0.5 \\
    Discount factor
    & 0.99 \\
    Reward clipping
    & [-1, 1] \\
    Optimizer
    & RMSProp \\
    \hspace{0.5cm} Weight decay
    & 0 \\
    \hspace{0.5cm} Smoothing constant
    & 0.99 \\
    \hspace{0.5cm} Momentum
    & 0 \\
    \hspace{0.5cm} epsilon
    & 0.01 \\
    Warmup steps
    & 0 \\
    Gradient norm clipping
    & 40 \\
    Scheduler
    & cosine \\
    \hspace{0.5cm} Steps between scheduler updates
    & 10,000 \\
    \hspace{0.5cm} Min learning rate
    & 0 \\ \\
    \end{tabular}
    \caption{Common hyperparameters for all experiments}
    \label{tab:common_params}
\end{table}{}

\newpage
\subsection{Hyperparameters For Stable Transformers}
\begin{table}[ht]
    \centering
    \begin{tabular}{l*{3}{c}r}
    \textbf{Parameter}
    & \textbf{Value}  \\
    \hline
    Embedding dimension
    & 256 \\
    No. of heads
    & 4 \\
    Dimension of head
    & 64 \\
    Position-wise feedforward size
    & 1024 \\
    Memory length
    & 100 \\
    Dropout
    & 0.1 \\ \\
    \end{tabular}
    \caption{Hyperparameters for Stable Transformer experiments}
    \label{tab:stable_params}
\end{table}{}

\subsection{Hyperparameters for Adaptive Transformers}
\begin{table}[ht]
    \centering
    \begin{tabular}{l*{3}{c}r}
    \textbf{Parameter}
    & \textbf{Value}  \\
    \hline
    Max attention span (memory length)
    & 400 \\
    Persistent memory
    & 0 \\
    Ramp length
    & 32 \\
    Initial value of adaptive span
    & 30\% of max span \\
    Dropout
    & 0.1 \\ \\
    \end{tabular}
    \caption{Hyperparameters for Adaptive experiments}
    \label{tab:adaptive_params}
\end{table}{}

\end{document}